\pdfoutput=1

\documentclass[11pt]{article}

\usepackage[preprint]{acl}

\usepackage{times}
\usepackage{latexsym}

\usepackage[T1]{fontenc}

\usepackage[utf8]{inputenc}

\usepackage{microtype}

\usepackage{inconsolata}

\usepackage{graphicx}

\usepackage{xurl}

\usepackage{float}

\usepackage{booktabs} 
\title{BCAmirs at SemEval-2024 Task 4: Beyond Words: A Multimodal and Multilingual Exploration of Persuasion in Memes 
}

\author{
  Amirhossein Abaskohi$^\dagger$\thanks{\hspace{0.4em}Equal Contribution}, Amirhossein Dabiriaghdam$^\ddagger$\footnotemark[1], Lele Wang$^\ddagger$, Giuseppe Carenini$^\dagger$ \\
  $^\dagger$Department of Computer Science, University of British Columbia \\
  $^\ddagger$Department of Electrical and Computer Engineering, University of British Columbia \\
  Vancouver, Canada \\
  \texttt{\{aabaskoh, carenini\}@cs.ubc.ca} \\
  \texttt{\{amirhossein, lelewang\}@ece.ubc.ca}
}

\begin{document}
\maketitle
\begin{abstract}
Memes, combining text and images, frequently use metaphors to convey persuasive messages, shaping public opinion. Motivated by this, our team engaged in SemEval-2024 Task 4, a hierarchical multi-label classification task designed to identify rhetorical and psychological persuasion techniques embedded within memes. To tackle this problem, we introduced a caption generation step to assess the modality gap and the impact of additional semantic information from images, which improved our result. Our best model utilizes GPT-4 generated captions alongside meme text to fine-tune RoBERTa as the text encoder and CLIP as the image encoder. It outperforms the baseline by a large margin in all 12 subtasks. In particular, it ranked in top-3 across all languages in Subtask 2a, and top-4 in Subtask 2b, demonstrating quantitatively strong performance. The improvement achieved by the introduced intermediate step is likely attributable to the metaphorical essence of images that challenges visual encoders. This highlights the potential for improving abstract visual semantics encoding.\footnote{This work is supported in part by the Institute for Computing, Information and Cognitive Systems (ICICS) at UBC.}\textsuperscript{,} 
\footnote{Our code is publicly available at \url{https://github.com/AmirAbaskohi/Beyond-Words-A-Multimodal-Exploration-of-Persuasion-in-Memes}.}

\end{abstract}

\section{Introduction}

In this digital age, the influence of persuasive techniques, particularly in memes, is a key focus. Propaganda, using various psychological techniques, shapes information for specific agendas. Research on political memes, such as \citet{kulkarni2017internet}'s work, emphasizes their role in communication and satire. 
Another study on COVID-19 memes \cite{wasike2022memes} underscores the importance of expert-sourced, objective memes in influencing public opinion and aiding public health campaigns. As a result, understanding persuasive techniques in memes within disinformation campaigns is crucial for grasping their impact on public perception and discourse. These campaigns usually succeed in influencing users by employing various rhetorical and psychological strategies in memes, including but not limited to \textit{causal oversimplification}, \textit{thought-terminating cliché}, and \textit{smear} techniques.

To address this concern, we participated in the SemEval-2024 shared task 4, as outlined by \cite{semeval2024task4}. The primary objective of this shared task is to develop models specifically designed to detect rhetorical and psychological techniques within memes. In summary, this task involves three subtasks. In Subtask 1 the input is the textual content of a meme only. This could include any written information present within the meme, and the goal is to identify one of the 20 persuasion techniques present in the meme's textual content. The identification is based on a hierarchical structure, and the techniques are organized in a tree-like fashion. Subtask 2a involves both textual and visual content analysis of memes, and information present in both the written content and the visual elements of the meme are considered. The task is to identify the presence of 22 persuasion techniques, utilizing a hierarchical structure similar to Subtask 1. Subtask 2b is a binary classification version of Subtask 2a. The training set released for all subtasks contains only English memes. However, alongside the English language, the test datasets contain memes in three low-resource languages (Arabic, Bulgarian, and North Macedonian) that aim to evaluate the zero-shot capability of the proposed models. 

\begin{table}
    \centering
    \small
    \begin{tabular}{lccc}
    \hline
    \textbf{Subtask} & \textbf{Ours} & \textbf{Baseline} & \textbf{Rank}\\
    \hline
    2a - English  & \textbf{70.497} & 44.706 & 3 / 15 \\
    2a - Bulgarian & \textbf{62.693} & 50.000 & 1 / 9\\
    2a - North Macedonian & \textbf{63.681} & 55.525 & 1 / 9 \\
    2a - Arabic & \textbf{52.613} & 48.649 & 1 / 7 \\
    \hline
    2b - English & \textbf{80.337} & 25.000 & 4 / 22\\ 
    2b - Bulgarian & \textbf{64.719} & 16.667 & 4 / 16\\ 
    2b - North Macedonian & \textbf{56.098} & 09.091 & 4 / 16\\ 
    2b - Arabic & \textbf{61.487} & 22.705 & 1 / 15 \\ 
    \hline
    1 - English & \textbf{69.857} &36.865 & 2 / 33 \\
    1 - Bulgarian & \textbf{44.834} & 28.377 & 13 / 20\\
    1 - North Macedonian & \textbf{39.298} & 30.692 & 12 / 20 \\
    1 - Arabic & \textbf{39.625} & 35.897 & 9 / 17\\
    \hline
    \end{tabular}
    \caption{Results of our best model (at the time of submitting evaluation results) on the test dataset of different subtasks. In the table the ranking and the values are based on hierarchical F1.}
    \label{tab:results-intro}
\vspace{-1em}
\end{table}

Although we participated in all subtasks, we specifically focused on Subtask 2 which uses both the textual and visual modality of memes to do a multi-label classification. To achieve better results, we introduced an intermediate step, the meme captioning step. Subsequently, we employed these generated captions to compare the performance of different models like LLaVA-1.5 \cite{liu2023llava}, Vicuna-1.5 \cite{vicuna2023}, BERT \cite{devlin2018bert}, and RoBERTa \cite{liu2019roberta}. This comparative analysis aimed to elucidate the role of the memes' text, the generated captions, and the memes' images in understanding persuasion techniques used in memes. The results of our best model (at the time of submitting evaluation results) that uses RoBERTa and our ranking relative to other teams in different subtasks are summarized in Table \ref{tab:results-intro}. It can be seen that our method performed well in the Subtask 2 (our main focus) for all four languages, and also the English subset of the first subtask. Our model struggled with non-English subsets of Subtask 1 since (I) we did not have access to the image of the meme and therefore no caption was available, and (II) our models only understood English, so we relied on a translation (using Google Translate\footnote{\url{https://translate.google.com}}) of the memes' text. 

Prior approaches have tried to narrow the gap between visual and textual realms to enhance image captioning. However, these methods primarily emphasized captioning visual details through textually enriched image features, rather than delving into the metaphorical significance inherent in images, particularly in the context of memes. To the best of our knowledge, this is the first work that focuses on the metaphorical semantic gap in multimodal language models to examine the gap between image and text modalities. Our ultimate goal was to gain insight into discrepancies between visual and textual metaphors in these systems. In summary, our contributions are twofold: (I) Addressing the classification problem of persuasion techniques in memes using multimodal models, and (II) Investigating the modality gap between textual and image components in multimodal models.


This paper is organized as follows: In Section \ref{sec:background}, we review prior research on hierarchical classification, persuasion techniques classification from memes, and the gap between textual and visual modalities. Section \ref{sec:method} introduces the datasets, discusses models, and outlines our approach for hierarchical persuasion technique classification. Section \ref{sec:experiments} presents and discusses our experiments and findings. Finally, in Section \ref{sec:conclusion}, we conclude our work, summarizing key contributions and suggesting directions for future research.

\section{Background}
\label{sec:background}

\paragraph{Modality Gap.}
In researching the modality gap between modalities, \citet{zhao2023chatbridge} presents ChatBridge, a novel multimodal large language model (MLLM) that employs language as a catalyst to bridge the gap between various modalities, such as text, image, video, and audio. By leveraging the expressive capabilities of language, ChatBridge connects different modalities using only language-paired bimodal data, showcasing strong quantitative and qualitative results on zero-shot multimodal tasks. Furthermore, \citet{chen2023lion} addresses the limitations of existing MLLMs in effectively extracting and reasoning visual knowledge. The proposed model, LION, injects dual-level visual knowledge, incorporating fine-grained spatial-aware visual knowledge and high-level semantic visual evidence. LION outperforms existing models in vision-language tasks, including image captioning, visual question answering, and visual grounding, through a two-stage training process. By extending these insights to the unique realm of memes, our work not only adds to the growing body of research on multimodal models but also sheds light on the gap between visual and textual modalities, especially in the metaphorical landscape.


\paragraph{Persuasion Technique Classification.}
In exploring persuasion techniques in texts and images, \citet{dimitrov-etal-2021-semeval} present a comprehensive framework for meme analysis. The study defines 22 techniques and provides an annotated dataset for conducting nuanced examinations of textual and multimodal memes. The incorporation of historical and mythological references adds depth to understanding the challenges in this domain. Moreover, in the work by \citet{messina-etal-2021-aimh}, the authors introduce transformer-based \cite{vaswani2017attention} models, VTTE and DVTT, for processing textual and visual content in memes. These models effectively identify persuasion techniques, with DVTT showing superior performance, particularly in fine-tuning feature extractors. Given the prevalence of Large Language Models (LLMs) employing similar architectures as DVTT, our experiments include utilizing LLMs and MLLMs, which we explain in Section \ref{sec:method}, to further investigate and advance the detection of persuasion techniques in memes.

\section{Methodology}
\label{sec:method}

Building upon prior research on MLLMs, the prevailing method involves tokenizing image concepts and conveying these tokens alongside textual tokens to a language model. While these models possess the ability to impart more semantic information from the image, their focus typically centers on identifying objects and their relationships within the image \cite{park2023refcap}. Consequently, this study explores the impact of initially prompting the model to generate descriptive information aimed at conveying semantic context. We utilize this information for data classification, comparing it to the conventional approach of fine-tuning an end-to-end model. In this section, we outline our approach for generating meme captions and subsequently discuss our classifiers.

\subsection{Caption Generation}
\label{subsec:caption-generation}

\begin{figure*}
    \centering
    \includegraphics[page=18,width=0.805\linewidth]{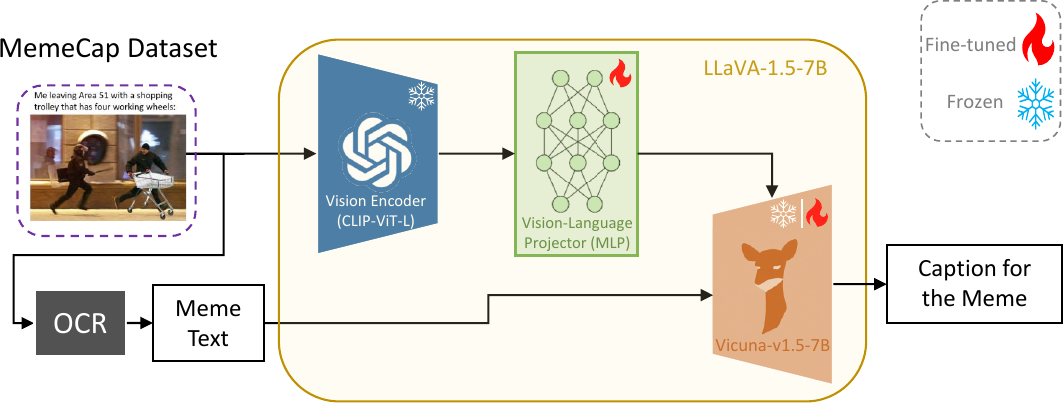}
    \caption{The figure depicts the supervised fine-tuning loop of the LLaVA-1.5-7B model on the MemeCap dataset for caption generation. The OCR module extracts text from the meme images. The vision encoder (CLIP), a frozen component of LLaVA-1.5-7B, processes the meme images. The vision-language projector bridges the gap between CLIP's representation and the embedding space of Vicuna. While CLIP remains frozen, the vision-language projector is fine-tuned. Vicuna component experimented with both frozen and fine-tuned setups to generate captions.}
    \label{fig:MemeCap}
\end{figure*}

In this paper we used three different models for generating meme captions: BLIP-2 \cite{li2023blip}, LLaVA-1.5-7B \cite{liu2023llava}, and GPT-4 \cite{openai2023gpt4}. We fine-tuned BLIP-2 and LLaVA-1.5 for generating captions and used GPT-4 in zero-shot settings. Based on our results which are explained in Appendix \ref{appendix:captioning-results}, we found out that LLaVA-1.5 outperforms BLIP-2 in the quality of generated captions. In order to fine-tune our meme captioning model, we used MemeCap \cite{hwang-shwartz-2023-memecap} dataset. Figure \ref{fig:MemeCap} illustrates the fine-tuning loop for LLaVA. This involved generating descriptions of memes capturing the conveyed message to uncover deeper layers of semantic understanding. Subsequently, we utilized this fine-tuned LLaVA to generate captions for the persuasion technique datasets. The generated captions provided supplementary data, offering further insights into the memes and enabling us to examine the effects of additional semantic information. Additionally, these captions were utilized to evaluate the modality gap within MLLMs. As elaborated in the subsequent section, we studied the distinctions between incorporating both the meme and its caption during the classification phase.



Considering our results in meme caption generation, discussed in Appendix \ref{appendix:captioning-results}, we identified two potential issues with the captions generated by our fine-tuned models. The first issue concerns the domain disparity between the MemeCap and persuasion datasets. The memes in the task's dataset often contain toxic content and usually require a deep understanding of background knowledge and events. The second issue relates to the brevity of captions in the MemeCap dataset, which typically only mentions the meme's final goal. This brevity may not provide sufficient information for detecting persuasion approaches. Consequently, we opted to generate captions using GPT-4 in zero-shot settings. GPT-4, the latest MLLM from OpenAI, was employed in our study utilizing the recently released API known as \textbf{gpt-4-vision-preview}. This model exhibits remarkable proficiency across diverse tasks such as visual question answering and image captioning. See Appendix \ref{appendix:prompt} for details in caption generation with GPT-4.

\subsection{Persuasion Technique Classification}
After generating captions for the memes, except for Subtask 1 where only the text written in the meme is provided, for classifying persuasion techniques we have three features available: meme, the text written in it, and our generated caption. In order to investigate the effect of our proposed model and assess the modality gap in MLLMs, we evaluate the effect of different combinations of these features. This will be explained further in Section \ref{sec:experiments}. As our classifier model, whether in a multi-label setting like subtasks 1 and 2a, or 2b which is binary classification, we used different families of models from LLMs, MLLMs, and Language Representation Models (LRMs). These models are as follows:

\paragraph{LLMs.}
Given the promising results of LLMs in semantic classification tasks \cite{sun-etal-2023-text, abaskohi-etal-2023-lm}, we opted to use them as our classifiers. To ensure a fair comparison and analyze the modality gap, we utilized the same LLM used in LLaVA, namely Vicuna. We initially fine-tuned the LLM solely with the text written in the meme, followed by fine-tuning with both the text in the meme and the meme's caption.

\paragraph{MLLMs.}
To assess the impact of employing our intermediate step of generating meme captions and using them alongside the memes, we required training an end-to-end model. To accomplish this, we fine-tuned LLaVA by incorporating the meme's image, the text written within memes with or without the captions associated with them. This augmentation aimed to leverage the additional information conveyed by the captions and potentially enhance the model's performance.

\paragraph{LRMs.}
In several of our experiments, we employed BERT and RoBERTa as our classifiers. These models utilize only the encoder component of the transformer architecture. Despite the growing dominance of LLMs in various benchmarks, these models remain highly potent, particularly in semantic-related tasks. We fine-tuned the large versions of these models as our classifiers, first solely with the text written in the meme and then with both the text in the meme and the meme's caption.

\paragraph{Multimodal LRMs.}

\begin{figure*}[!ht]
    \centering
    \includegraphics[page=17,width=0.88\linewidth]{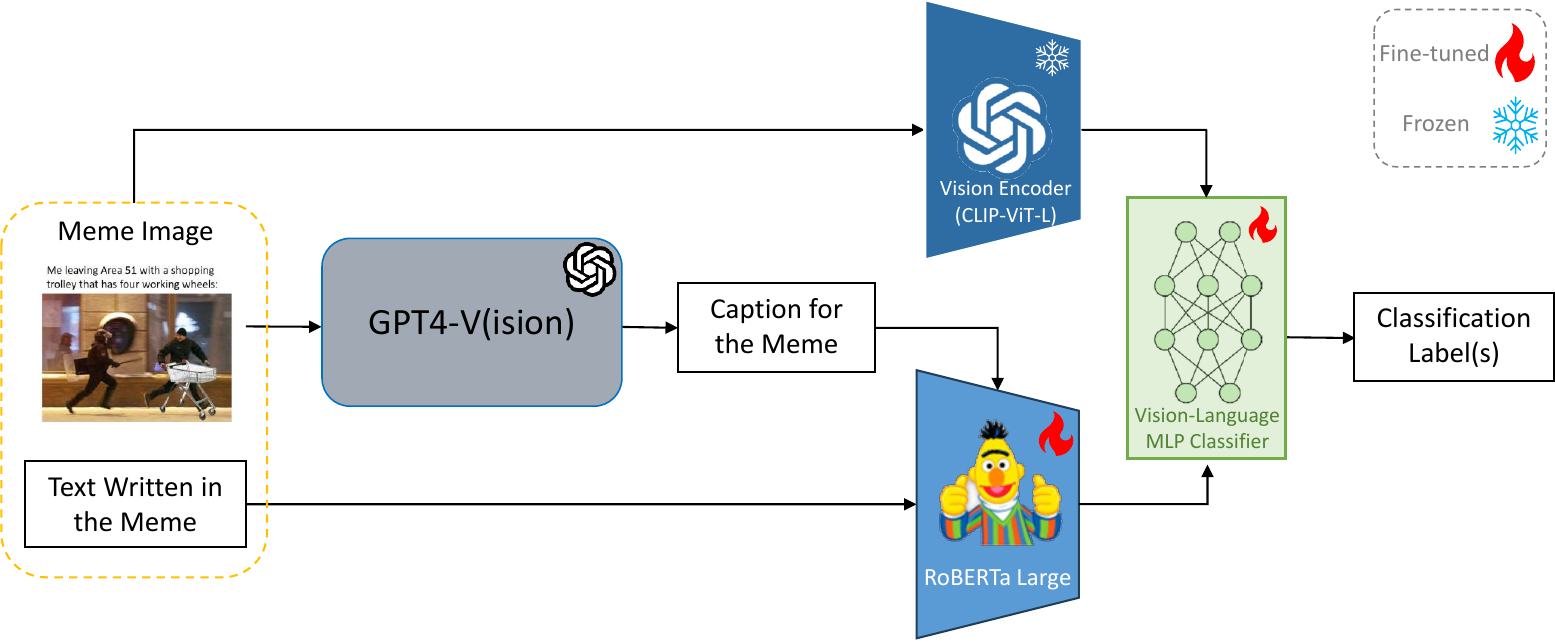}
    \caption{The figure illustrates the architecture of ConcatRoBERTa, our best-performing model. The GPT4-V(ision) component generates a descriptive caption of the meme image. The caption is then combined with the text written in the meme, which is processed by the RoBERTa. The Vision encoder utilizes a pre-trained vision transformer model (CLIP-ViT), to encode and analyze the visual elements of the meme. The MLP Classifier takes the combined visual and textual representations and classifies the meme. RoBERTa and the MLP classifiers are fine-tuned, while CLIP remains frozen.}
    \label{fig:concatRoBERTa/CLIP}
\end{figure*}

After exploring the impact of MLMMs alongside LLMs, we delved into a BERT variant with visual understanding capabilities. Initially, we fine-tuned VisualBERT \cite{li1908visualbert} on both memes and their accompanying text, with and without captions. However, given the unavailability of a pre-trained large version of VisualBERT we devised a concatenated model comprising RoBERTa-large and a vision encoder, inspired by an example from \cite{singh2020mmf}. We experimented with CLIP-ViT-large \cite{radford2021learning} as the component for the vision encoder. Subsequently, we concatenated the encoded features from CLIP with encoded features from RoBERTa and employed a linear classifier to determine the class based on the encoded information. We call this model \textit{ConcatRoBERTa} (see Figure \ref{fig:concatRoBERTa/CLIP}). To maintain consistency with previous MLLM-based methods, CLIP were frozen during the training phase. 



\section{Experiments}
\label{sec:experiments}

In this section, we outline our conducted experiments and provide a concise overview of the results obtained. Readers are referred to Appendix \ref{Appendix:eval-setting} for the details of our experimental settings. 

\begin{table*}[!ht]  
  \centering
  \begin{tabular}{lccc}
    \toprule
    \textbf{Model} & \textbf{H-F1} & \textbf{H-Precision} & \textbf{H-Recall} \\
    \midrule		
    LLaVA-1.5 (image) & 58.21 & 62.74 & 54.31 \\
    LLaVA-1.5 (image+text) & 62.59 & 66.00 & 59.51 \\
    LLaVA-1.5 (image+text+caption from LLaVA-1.5) & 63.33 & 67.02 & 60.02 \\ 
    \midrule		
    Vicuna-1.5 (text) & 62.69 & 71.03 & 56.10 \\
    Vicuna-1.5 (text+caption from LLaVA-1.5) & 63.11 & 70.86 & 56.88 \\
    Vicuna-1.5 (text+caption from GPT-4) & 65.337 & 75.204 & 57.759\\
    \midrule		
    BERT (text) & 64.881 & 75.400 & 56.938 \\
    BERT (text+caption from LLaVA-1.5) & 66.455 & 74.229 & 60.155 \\
    BERT (text+caption from GPT-4) & 66.829  & 75.958 & 59.659 \\
    \midrule		
    RoBERTa (text) & 66.740 & \underline{76.846} & 58.983 \\
    RoBERTa (text+caption from LLaVA-1.5) & 67.750 & 73.699 & 62.690 \\
    RoBERTa (text+caption from GPT-4) & \underline{69.913} & \textbf{76.999} & 64.021 \\
    \midrule		
    VisualBERT (image+text) & 51.496 & 39.779 & \textbf{72.998} \\
    VisualBERT (image+text+caption from LLaVA-1.5) & 57.714 & 57.841 & 62.690 \\
    \midrule		
    ConcatRoBERTa (image+text) & 65.188 & 73.443 & 58.601 \\
    ConcatRoBERTa (image+text+caption from LLaVA-1.5) & 67.166 & 75.283 & 60.629 \\
    ConcatRoBERTa (image+text+caption from GPT-4) & \textbf{71.115} & 76.101 & \underline{66.742} \\
    \midrule		
    Baseline & 44.706 & 68.778 & 33.116 \\
    \bottomrule
  \end{tabular}
\caption{Comparison of results of different methods on dev set of Subtask 2a. H-F1, H-Precision, and H-Recall, are hierarchical-F1, -precision, and -recall respectively. As expected, models prefer to receive more information about the image, and models incorporating all features (e.g., text, caption, and image) tend to perform better. However, captions appear to be more informative. This suggests that although some information from the image may not be fully conveyed through text, utilizing models to initially analyze the image, particularly in meme tasks like this, and then prompting them to make decisions based on that analysis, yields superior performance compared to making decisions without leveraging their full capabilities.}
  \label{table:2a-dev}
\end{table*}

For evaluation of the performance of the models for hierarchical classifications, we use hierarchical-precision, -recall, and -F1 introduced by \citet{kiritchenko2006learning}. For more information about hierarchical evaluation metrics, see Appendix \ref{appendix:Hierarchical}. 

In the initial set of experiments, we perform hierarchical multilabel classification using the textual content of memes to fine-tune unimodal models (Vicuna, BERT, and RoBERTa) directly for identifying specific persuasion techniques. We compare this approach to multimodal models (LLaVA, VisualBERT, and ConcatRoBERTa) where both textual and visual contents of the meme are provided. Additionally, we conduct a similar experiment with LLaVA model, feeding only the image of the meme without textual data. This comparison aims to assess the information derived from each modality. The motivation behind our decision to compare encoder-only LRMs such as RoBERTa or VisualBERT to larger generative models like Vicuna or LLaVA, is their promising results in classification tasks compared to generative models \cite{sarrouti-etal-2022-comparing, jiang-etal-2023-brainteaser}.  
During the second stage of our experiments, we proposed to create captions for the memes, and subsequently augment the original data with these generated captions. 
This additional step aims to capture more information from the meme image and adopt this additional data to improve the results of the hierarchical classification of the memes. In this phase, we mainly focused on subtask 2a. The results of our different methods on the dev set of subtask 2a are presented in Table \ref{table:2a-dev}. 

From Table \ref{table:2a-dev}, we can see the best performing model is ConcatRoBERTa, which has both the image and the text written on the meme as well as the caption generated by GPT-4 as its inputs (Figure \ref{fig:concatRoBERTa/CLIP}) with hierarchical F1 score of 71.115 on the dev set of subtask2a. It is worth mentioning that due to time constraints, we could not evaluate test datasets using ConcatRoBERTa by the evaluation deadline, therefore, the submitted results for the test dataset in Table \ref{tab:results-intro} are from RoBERTa model (our second best model). It might be unexpected that MLLMs like LLaVA with text and image of the meme as their input do not perform as well as LLMs like Vicuna with text and caption of the meme in this particular task. This discrepancy could be attributed to the metaphorical nature of memes. Vision encoders, such as CLIP, are primarily trained to comprehend the visual aspects of an image, lacking a focus on the metaphorical meanings embedded in those visual elements. In contrast, language models are more adept at understanding metaphors, given their greater exposure to such linguistic nuances in textual data which has been shown previously \cite{hwang-shwartz-2023-memecap}. Note the improvement in the results when employing GPT-4 for caption generation instead of LLaVA. As mentioned earlier, it is due to the domain disparity between MemeCap and this task's dataset. Regarding the superiority of the results of fine-tuned LRMs such as RoBERTa compared to LLMs like Vicuna, we argue that LLMs in general tasks are better but often for certain tasks a well-implemented LRM can outperform LLMs. In other words, the performance of LLMs fluctuates significantly based on the limitations of the data and the specific application context. This observation can be attributed to the use of a relatively small generative language model (with only 7B parameters) for a challenging task. Finally, it is not surprising that VisualBERT's results are not as good as other larger models since we only had access to the base version of pre-trained VisualBERT.


Another observation is that by adding an intermediate step of caption generation, results are improved when it is used in a supervised learning manner. In contrast, for the in-context learning scheme (Appendix \ref{appendix:in-context-learning}), we note that the additional information extracted from memes, specifically captions, did not improve but rather worsened the results. The diverse nature of meme captions, including more details compared to the text within the memes, may misguide the model in focusing on relevant features. In such a setting, the models' in-context learning ability is limited, and giving more information only confuses the model without any gain. Even we tried to use GPT-4 (in a zero-shot setting) for subtask 2b, and its results on the dev set were comparable but worse than using our proposed method (RoBERTa with generated caption from GPT-4), i.e., 73.242 and 79.667 versus 78.434 and 81.333 for macro- and micro-F1, respectively.
  
\section{Conclusion and Future Work}
\label{sec:conclusion}


This paper explores the persuasive communication within memes, emphasizing their role in shaping public perception. Through participation in the SemEval-2024 shared task 4, our study delves into the detection of rhetorical and psychological techniques within memes. By employing multimodal models and introducing an intermediate step of meme captioning using LLaVA and GPT-4, we aimed to bridge the gap between textual and visual modalities, thereby enhancing the classification of persuasion techniques. Our experiments demonstrate the effectiveness of this approach, with our best model, ConcatRoBERTa, achieving notable performance improvements. However, we observed that the performance gains varied based on the dataset's nature and the models' sophistication. Nevertheless, our findings contribute to advancing understanding in this domain and pave the way for future research endeavors aimed at combating online disinformation campaigns.  

Regarding future work, a deeper analysis into why the model struggles to utilize its image analysis capabilities for classification, despite its proficiency in generating captions (even in zero-shot settings with GPT-4), could be explored through the implementation of chain-of-thought approaches. Additionally, exploring how well the proposed method withstands adversarial attacks is another interesting direction. Adversarial examples, as shown by different studies \cite{moosavi2016deepfool, sadrizadeh2023targeted, zhao2024evaluating}, have uncovered vulnerabilities in neural models across various tasks. Studying how adding the caption generation step affects the adversarial robustness of our approach compared to end-to-end methods for this task holds promise for future research.


\bibliography{anthology}

\begin{thebibliography}{35}
\expandafter\ifx\csname natexlab\endcsname\relax\def\natexlab#1{#1}\fi

\bibitem[{Abaskohi et~al.(2023)Abaskohi, Rothe, and Yaghoobzadeh}]{abaskohi-etal-2023-lm}
Amirhossein Abaskohi, Sascha Rothe, and Yadollah Yaghoobzadeh. 2023.
\newblock \href {https://doi.org/10.18653/v1/2023.acl-short.59} {{LM}-{CPPF}: Paraphrasing-guided data augmentation for contrastive prompt-based few-shot fine-tuning}.
\newblock In \emph{Proceedings of the 61st Annual Meeting of the Association for Computational Linguistics (Volume 2: Short Papers)}, pages 670--681, Toronto, Canada. Association for Computational Linguistics.

\bibitem[{Chen et~al.(2023)Chen, Shen, Shao, Deng, and Nie}]{chen2023lion}
Gongwei Chen, Leyang Shen, Rui Shao, Xiang Deng, and Liqiang Nie. 2023.
\newblock Lion: Empowering multimodal large language model with dual-level visual knowledge.
\newblock \emph{arXiv preprint arXiv:2311.11860}.

\bibitem[{Chiang et~al.(2023)Chiang, Li, Lin, Sheng, Wu, Zhang, Zheng, Zhuang, Zhuang, Gonzalez, Stoica, and Xing}]{vicuna2023}
Wei-Lin Chiang, Zhuohan Li, Zi~Lin, Ying Sheng, Zhanghao Wu, Hao Zhang, Lianmin Zheng, Siyuan Zhuang, Yonghao Zhuang, Joseph~E. Gonzalez, Ion Stoica, and Eric~P. Xing. 2023.
\newblock \href {https://lmsys.org/blog/2023-03-30-vicuna/} {Vicuna: An open-source chatbot impressing gpt-4 with 90\%* chatgpt quality}.

\bibitem[{Devlin et~al.(2018)Devlin, Chang, Lee, and Toutanova}]{devlin2018bert}
Jacob Devlin, Ming-Wei Chang, Kenton Lee, and Kristina Toutanova. 2018.
\newblock Bert: Pre-training of deep bidirectional transformers for language understanding.
\newblock \emph{arXiv preprint arXiv:1810.04805}.

\bibitem[{Dimitrov et~al.(2024)Dimitrov, Alam, Hasanain, Hasnat, Silvestri, Nakov, and Da~San~Martino}]{semeval2024task4}
Dimitar Dimitrov, Firoj Alam, Maram Hasanain, Abul Hasnat, Fabrizio Silvestri, Preslav Nakov, and Giovanni Da~San~Martino. 2024.
\newblock Semeval-2024 task 4: Multilingual detection of persuasion techniques in memes.
\newblock In \emph{Proceedings of the 18th International Workshop on Semantic Evaluation}, SemEval 2024, Mexico City, Mexico.

\bibitem[{Dimitrov et~al.(2021)Dimitrov, Bin~Ali, Shaar, Alam, Silvestri, Firooz, Nakov, and Da~San~Martino}]{dimitrov-etal-2021-semeval}
Dimitar Dimitrov, Bishr Bin~Ali, Shaden Shaar, Firoj Alam, Fabrizio Silvestri, Hamed Firooz, Preslav Nakov, and Giovanni Da~San~Martino. 2021.
\newblock \href {https://doi.org/10.18653/v1/2021.semeval-1.7} {{S}em{E}val-2021 task 6: Detection of persuasion techniques in texts and images}.
\newblock In \emph{Proceedings of the 15th International Workshop on Semantic Evaluation (SemEval-2021)}, pages 70--98, Online. Association for Computational Linguistics.

\bibitem[{He et~al.(2021)He, Liu, Gao, and Chen}]{he2021deberta}
Pengcheng He, Xiaodong Liu, Jianfeng Gao, and Weizhu Chen. 2021.
\newblock \href {https://openreview.net/forum?id=XPZIaotutsD} {Deberta: Decoding-enhanced bert with disentangled attention}.
\newblock In \emph{International Conference on Learning Representations}.

\bibitem[{Hu et~al.(2021)Hu, Shen, Wallis, Allen-Zhu, Li, Wang, Wang, and Chen}]{hu2021lora}
Edward~J Hu, Yelong Shen, Phillip Wallis, Zeyuan Allen-Zhu, Yuanzhi Li, Shean Wang, Lu~Wang, and Weizhu Chen. 2021.
\newblock Lora: Low-rank adaptation of large language models.
\newblock \emph{arXiv preprint arXiv:2106.09685}.

\bibitem[{Hwang and Shwartz(2023)}]{hwang-shwartz-2023-memecap}
EunJeong Hwang and Vered Shwartz. 2023.
\newblock \href {https://doi.org/10.18653/v1/2023.emnlp-main.89} {{M}eme{C}ap: A dataset for captioning and interpreting memes}.
\newblock In \emph{Proceedings of the 2023 Conference on Empirical Methods in Natural Language Processing}, pages 1433--1445, Singapore. Association for Computational Linguistics.

\bibitem[{Jiang et~al.(2023)Jiang, Ilievski, Ma, and Sourati}]{jiang-etal-2023-brainteaser}
Yifan Jiang, Filip Ilievski, Kaixin Ma, and Zhivar Sourati. 2023.
\newblock \href {https://doi.org/10.18653/v1/2023.emnlp-main.885} {{BRAINTEASER}: Lateral thinking puzzles for large language models}.
\newblock In \emph{Proceedings of the 2023 Conference on Empirical Methods in Natural Language Processing}, pages 14317--14332, Singapore. Association for Computational Linguistics.

\bibitem[{Kingma and Ba(2014)}]{kingma2014adam}
Diederik~P Kingma and Jimmy Ba. 2014.
\newblock Adam: A method for stochastic optimization.
\newblock \emph{arXiv preprint arXiv:1412.6980}.

\bibitem[{Kiritchenko et~al.(2006)Kiritchenko, Matwin, Nock, and Famili}]{kiritchenko2006learning}
Svetlana Kiritchenko, Stan Matwin, Richard Nock, and A~Fazel Famili. 2006.
\newblock Learning and evaluation in the presence of class hierarchies: Application to text categorization.
\newblock In \emph{Advances in Artificial Intelligence: 19th Conference of the Canadian Society for Computational Studies of Intelligence, Canadian AI 2006, Qu{\'e}bec City, Qu{\'e}bec, Canada, June 7-9, 2006. Proceedings 19}, pages 395--406. Springer.

\bibitem[{Kulkarni(2017)}]{kulkarni2017internet}
Anushka Kulkarni. 2017.
\newblock Internet meme and political discourse: A study on the impact of internet meme as a tool in communicating political satire.
\newblock \emph{Journal of Content, Community \& Communication Amity School of Communication}, 6.

\bibitem[{Li et~al.(2023)Li, Li, Savarese, and Hoi}]{li2023blip}
Junnan Li, Dongxu Li, Silvio Savarese, and Steven Hoi. 2023.
\newblock Blip-2: Bootstrapping language-image pre-training with frozen image encoders and large language models.
\newblock \emph{arXiv preprint arXiv:2301.12597}.

\bibitem[{Li et~al.()Li, Yatskar, Yin, Hsieh, and Chang}]{li1908visualbert}
Liunian~Harold Li, Mark Yatskar, D~Yin, CJ~Hsieh, and KW~Chang.
\newblock Visualbert: A simple and performant baseline for vision and language. arxiv 2019.
\newblock \emph{arXiv preprint arXiv:1908.03557}.

\bibitem[{Lin(2004)}]{lin2004rouge}
Chin-Yew Lin. 2004.
\newblock Rouge: A package for automatic evaluation of summaries.
\newblock In \emph{Text summarization branches out}, pages 74--81.

\bibitem[{Liu et~al.(2023)Liu, Li, Wu, and Lee}]{liu2023llava}
Haotian Liu, Chunyuan Li, Qingyang Wu, and Yong~Jae Lee. 2023.
\newblock Visual instruction tuning.
\newblock In \emph{NeurIPS}.

\bibitem[{Liu et~al.(2019)Liu, Ott, Goyal, Du, Joshi, Chen, Levy, Lewis, Zettlemoyer, and Stoyanov}]{liu2019roberta}
Yinhan Liu, Myle Ott, Naman Goyal, Jingfei Du, Mandar Joshi, Danqi Chen, Omer Levy, Mike Lewis, Luke Zettlemoyer, and Veselin Stoyanov. 2019.
\newblock Roberta: A robustly optimized bert pretraining approach.
\newblock \emph{arXiv preprint arXiv:1907.11692}.

\bibitem[{Mangrulkar et~al.(2022)Mangrulkar, Gugger, Debut, Belkada, Paul, and Bossan}]{peft}
Sourab Mangrulkar, Sylvain Gugger, Lysandre Debut, Younes Belkada, Sayak Paul, and Benjamin Bossan. 2022.
\newblock Peft: State-of-the-art parameter-efficient fine-tuning methods.
\newblock \url{https://github.com/huggingface/peft}.

\bibitem[{Messina et~al.(2021)Messina, Falchi, Gennaro, and Amato}]{messina-etal-2021-aimh}
Nicola Messina, Fabrizio Falchi, Claudio Gennaro, and Giuseppe Amato. 2021.
\newblock \href {https://doi.org/10.18653/v1/2021.semeval-1.140} {{AIMH} at {S}em{E}val-2021 task 6: Multimodal classification using an ensemble of transformer models}.
\newblock In \emph{Proceedings of the 15th International Workshop on Semantic Evaluation (SemEval-2021)}, pages 1020--1026, Online. Association for Computational Linguistics.

\bibitem[{Moosavi-Dezfooli et~al.(2016)Moosavi-Dezfooli, Fawzi, and Frossard}]{moosavi2016deepfool}
Seyed-Mohsen Moosavi-Dezfooli, Alhussein Fawzi, and Pascal Frossard. 2016.
\newblock Deepfool: a simple and accurate method to fool deep neural networks.
\newblock In \emph{Proceedings of the IEEE conference on computer vision and pattern recognition}, pages 2574--2582.

\bibitem[{OpenAI et~al.(2023)OpenAI, :, Achiam, Adler, Agarwal, Ahmad, Akkaya, Aleman, Almeida, Altenschmidt, Altman, Anadkat, Avila, Babuschkin, Balaji, Balcom, Baltescu, Bao, Bavarian, Belgum, Bello, Berdine, Bernadett-Shapiro, Berner, Bogdonoff, Boiko, Boyd, Brakman, Brockman, Brooks, Brundage, Button, Cai, Campbell, Cann, Carey, Carlson, Carmichael, Chan, Chang, Chantzis, Chen, Chen, Chen, Chen, Chen, Chess, Cho, Chu, Chung, Cummings, Currier, Dai, Decareaux, Degry, Deutsch, Deville, Dhar, Dohan, Dowling, Dunning, Ecoffet, Eleti, Eloundou, Farhi, Fedus, Felix, Fishman, Forte, Fulford, Gao, Georges, Gibson, Goel, Gogineni, Goh, Gontijo-Lopes, Gordon, Grafstein, Gray, Greene, Gross, Gu, Guo, Hallacy, Han, Harris, He, Heaton, Heidecke, Hesse, Hickey, Hickey, Hoeschele, Houghton, Hsu, Hu, Hu, Huizinga, Jain, Jain, Jang, Jiang, Jiang, Jin, Jin, Jomoto, Jonn, Jun, Kaftan, Łukasz Kaiser, Kamali, Kanitscheider, Keskar, Khan, Kilpatrick, Kim, Kim, Kim, Kirchner, Kiros, Knight, Kokotajlo, Łukasz Kondraciuk,
  Kondrich, Konstantinidis, Kosic, Krueger, Kuo, Lampe, Lan, Lee, Leike, Leung, Levy, Li, Lim, Lin, Lin, Litwin, Lopez, Lowe, Lue, Makanju, Malfacini, Manning, Markov, Markovski, Martin, Mayer, Mayne, McGrew, McKinney, McLeavey, McMillan, McNeil, Medina, Mehta, Menick, Metz, Mishchenko, Mishkin, Monaco, Morikawa, Mossing, Mu, Murati, Murk, Mély, Nair, Nakano, Nayak, Neelakantan, Ngo, Noh, Ouyang, O'Keefe, Pachocki, Paino, Palermo, Pantuliano, Parascandolo, Parish, Parparita, Passos, Pavlov, Peng, Perelman, de~Avila Belbute~Peres, Petrov, de~Oliveira~Pinto, Michael, Pokorny, Pokrass, Pong, Powell, Power, Power, Proehl, Puri, Radford, Rae, Ramesh, Raymond, Real, Rimbach, Ross, Rotsted, Roussez, Ryder, Saltarelli, Sanders, Santurkar, Sastry, Schmidt, Schnurr, Schulman, Selsam, Sheppard, Sherbakov, Shieh, Shoker, Shyam, Sidor, Sigler, Simens, Sitkin, Slama, Sohl, Sokolowsky, Song, Staudacher, Such, Summers, Sutskever, Tang, Tezak, Thompson, Tillet, Tootoonchian, Tseng, Tuggle, Turley, Tworek, Uribe, Vallone,
  Vijayvergiya, Voss, Wainwright, Wang, Wang, Wang, Ward, Wei, Weinmann, Welihinda, Welinder, Weng, Weng, Wiethoff, Willner, Winter, Wolrich, Wong, Workman, Wu, Wu, Wu, Xiao, Xu, Yoo, Yu, Yuan, Zaremba, Zellers, Zhang, Zhang, Zhao, Zheng, Zhuang, Zhuk, and Zoph}]{openai2023gpt4}
OpenAI, :, Josh Achiam, Steven Adler, Sandhini Agarwal, Lama Ahmad, Ilge Akkaya, Florencia~Leoni Aleman, Diogo Almeida, Janko Altenschmidt, Sam Altman, Shyamal Anadkat, Red Avila, Igor Babuschkin, Suchir Balaji, Valerie Balcom, Paul Baltescu, Haiming Bao, Mo~Bavarian, Jeff Belgum, Irwan Bello, Jake Berdine, Gabriel Bernadett-Shapiro, Christopher Berner, Lenny Bogdonoff, Oleg Boiko, Madelaine Boyd, Anna-Luisa Brakman, Greg Brockman, Tim Brooks, Miles Brundage, Kevin Button, Trevor Cai, Rosie Campbell, Andrew Cann, Brittany Carey, Chelsea Carlson, Rory Carmichael, Brooke Chan, Che Chang, Fotis Chantzis, Derek Chen, Sully Chen, Ruby Chen, Jason Chen, Mark Chen, Ben Chess, Chester Cho, Casey Chu, Hyung~Won Chung, Dave Cummings, Jeremiah Currier, Yunxing Dai, Cory Decareaux, Thomas Degry, Noah Deutsch, Damien Deville, Arka Dhar, David Dohan, Steve Dowling, Sheila Dunning, Adrien Ecoffet, Atty Eleti, Tyna Eloundou, David Farhi, Liam Fedus, Niko Felix, Simón~Posada Fishman, Juston Forte, Isabella Fulford, Leo Gao,
  Elie Georges, Christian Gibson, Vik Goel, Tarun Gogineni, Gabriel Goh, Rapha Gontijo-Lopes, Jonathan Gordon, Morgan Grafstein, Scott Gray, Ryan Greene, Joshua Gross, Shixiang~Shane Gu, Yufei Guo, Chris Hallacy, Jesse Han, Jeff Harris, Yuchen He, Mike Heaton, Johannes Heidecke, Chris Hesse, Alan Hickey, Wade Hickey, Peter Hoeschele, Brandon Houghton, Kenny Hsu, Shengli Hu, Xin Hu, Joost Huizinga, Shantanu Jain, Shawn Jain, Joanne Jang, Angela Jiang, Roger Jiang, Haozhun Jin, Denny Jin, Shino Jomoto, Billie Jonn, Heewoo Jun, Tomer Kaftan, Łukasz Kaiser, Ali Kamali, Ingmar Kanitscheider, Nitish~Shirish Keskar, Tabarak Khan, Logan Kilpatrick, Jong~Wook Kim, Christina Kim, Yongjik Kim, Hendrik Kirchner, Jamie Kiros, Matt Knight, Daniel Kokotajlo, Łukasz Kondraciuk, Andrew Kondrich, Aris Konstantinidis, Kyle Kosic, Gretchen Krueger, Vishal Kuo, Michael Lampe, Ikai Lan, Teddy Lee, Jan Leike, Jade Leung, Daniel Levy, Chak~Ming Li, Rachel Lim, Molly Lin, Stephanie Lin, Mateusz Litwin, Theresa Lopez, Ryan Lowe,
  Patricia Lue, Anna Makanju, Kim Malfacini, Sam Manning, Todor Markov, Yaniv Markovski, Bianca Martin, Katie Mayer, Andrew Mayne, Bob McGrew, Scott~Mayer McKinney, Christine McLeavey, Paul McMillan, Jake McNeil, David Medina, Aalok Mehta, Jacob Menick, Luke Metz, Andrey Mishchenko, Pamela Mishkin, Vinnie Monaco, Evan Morikawa, Daniel Mossing, Tong Mu, Mira Murati, Oleg Murk, David Mély, Ashvin Nair, Reiichiro Nakano, Rajeev Nayak, Arvind Neelakantan, Richard Ngo, Hyeonwoo Noh, Long Ouyang, Cullen O'Keefe, Jakub Pachocki, Alex Paino, Joe Palermo, Ashley Pantuliano, Giambattista Parascandolo, Joel Parish, Emy Parparita, Alex Passos, Mikhail Pavlov, Andrew Peng, Adam Perelman, Filipe de~Avila Belbute~Peres, Michael Petrov, Henrique~Ponde de~Oliveira~Pinto, Michael, Pokorny, Michelle Pokrass, Vitchyr Pong, Tolly Powell, Alethea Power, Boris Power, Elizabeth Proehl, Raul Puri, Alec Radford, Jack Rae, Aditya Ramesh, Cameron Raymond, Francis Real, Kendra Rimbach, Carl Ross, Bob Rotsted, Henri Roussez, Nick Ryder,
  Mario Saltarelli, Ted Sanders, Shibani Santurkar, Girish Sastry, Heather Schmidt, David Schnurr, John Schulman, Daniel Selsam, Kyla Sheppard, Toki Sherbakov, Jessica Shieh, Sarah Shoker, Pranav Shyam, Szymon Sidor, Eric Sigler, Maddie Simens, Jordan Sitkin, Katarina Slama, Ian Sohl, Benjamin Sokolowsky, Yang Song, Natalie Staudacher, Felipe~Petroski Such, Natalie Summers, Ilya Sutskever, Jie Tang, Nikolas Tezak, Madeleine Thompson, Phil Tillet, Amin Tootoonchian, Elizabeth Tseng, Preston Tuggle, Nick Turley, Jerry Tworek, Juan Felipe~Cerón Uribe, Andrea Vallone, Arun Vijayvergiya, Chelsea Voss, Carroll Wainwright, Justin~Jay Wang, Alvin Wang, Ben Wang, Jonathan Ward, Jason Wei, CJ~Weinmann, Akila Welihinda, Peter Welinder, Jiayi Weng, Lilian Weng, Matt Wiethoff, Dave Willner, Clemens Winter, Samuel Wolrich, Hannah Wong, Lauren Workman, Sherwin Wu, Jeff Wu, Michael Wu, Kai Xiao, Tao Xu, Sarah Yoo, Kevin Yu, Qiming Yuan, Wojciech Zaremba, Rowan Zellers, Chong Zhang, Marvin Zhang, Shengjia Zhao, Tianhao
  Zheng, Juntang Zhuang, William Zhuk, and Barret Zoph. 2023.
\newblock \href {http://arxiv.org/abs/2303.08774} {Gpt-4 technical report}.

\bibitem[{Park and Paik(2023)}]{park2023refcap}
Seokmok Park and Joonki Paik. 2023.
\newblock Refcap: image captioning with referent objects attributes.
\newblock \emph{Scientific Reports}, 13(1):21577.

\bibitem[{Post(2018)}]{post-2018-call}
Matt Post. 2018.
\newblock \href {https://www.aclweb.org/anthology/W18-6319} {A call for clarity in reporting {BLEU} scores}.
\newblock In \emph{Proceedings of the Third Conference on Machine Translation: Research Papers}, pages 186--191, Belgium, Brussels. Association for Computational Linguistics.

\bibitem[{Radford et~al.(2021)Radford, Kim, Hallacy, Ramesh, Goh, Agarwal, Sastry, Askell, Mishkin, Clark et~al.}]{radford2021learning}
Alec Radford, Jong~Wook Kim, Chris Hallacy, Aditya Ramesh, Gabriel Goh, Sandhini Agarwal, Girish Sastry, Amanda Askell, Pamela Mishkin, Jack Clark, et~al. 2021.
\newblock Learning transferable visual models from natural language supervision.
\newblock In \emph{International conference on machine learning}, pages 8748--8763. PMLR.

\bibitem[{Sadrizadeh et~al.(2023)Sadrizadeh, Aghdam, Dolamic, and Frossard}]{sadrizadeh2023targeted}
Sahar Sadrizadeh, AmirHossein~Dabiri Aghdam, Ljiljana Dolamic, and Pascal Frossard. 2023.
\newblock Targeted adversarial attacks against neural machine translation.
\newblock In \emph{ICASSP 2023-2023 IEEE International Conference on Acoustics, Speech and Signal Processing (ICASSP)}, pages 1--5. IEEE.

\bibitem[{Sarrouti et~al.(2022)Sarrouti, Tao, and Mamy~Randriamihaja}]{sarrouti-etal-2022-comparing}
Mourad Sarrouti, Carson Tao, and Yoann Mamy~Randriamihaja. 2022.
\newblock \href {https://doi.org/10.18653/v1/2022.bionlp-1.37} {Comparing encoder-only and encoder-decoder transformers for relation extraction from biomedical texts: An empirical study on ten benchmark datasets}.
\newblock In \emph{Proceedings of the 21st Workshop on Biomedical Language Processing}, pages 376--382, Dublin, Ireland. Association for Computational Linguistics.

\bibitem[{Singh et~al.(2020)Singh, Goswami, Natarajan, Jiang, Chen, Shah, Rohrbach, Batra, and Parikh}]{singh2020mmf}
Amanpreet Singh, Vedanuj Goswami, Vivek Natarajan, Yu~Jiang, Xinlei Chen, Meet Shah, Marcus Rohrbach, Dhruv Batra, and Devi Parikh. 2020.
\newblock Mmf: A multimodal framework for vision and language research.
\newblock \url{https://github.com/facebookresearch/mmf}.

\bibitem[{Sun et~al.(2023)Sun, Li, Li, Wu, Guo, Zhang, and Wang}]{sun-etal-2023-text}
Xiaofei Sun, Xiaoya Li, Jiwei Li, Fei Wu, Shangwei Guo, Tianwei Zhang, and Guoyin Wang. 2023.
\newblock \href {https://doi.org/10.18653/v1/2023.findings-emnlp.603} {Text classification via large language models}.
\newblock In \emph{Findings of the Association for Computational Linguistics: EMNLP 2023}, pages 8990--9005, Singapore. Association for Computational Linguistics.

\bibitem[{Vaswani et~al.(2017)Vaswani, Shazeer, Parmar, Uszkoreit, Jones, Gomez, Kaiser, and Polosukhin}]{vaswani2017attention}
Ashish Vaswani, Noam Shazeer, Niki Parmar, Jakob Uszkoreit, Llion Jones, Aidan~N Gomez, {\L}ukasz Kaiser, and Illia Polosukhin. 2017.
\newblock Attention is all you need.
\newblock \emph{Advances in neural information processing systems}, 30.

\bibitem[{Wasike(2022)}]{wasike2022memes}
Ben Wasike. 2022.
\newblock Memes, memes, everywhere, nor any meme to trust: Examining the credibility and persuasiveness of covid-19-related memes.
\newblock \emph{Journal of Computer-Mediated Communication}, 27(2):zmab024.

\bibitem[{Zhang et~al.(2022)Zhang, Roller, Goyal, Artetxe, Chen, Chen, Dewan, Diab, Li, Lin et~al.}]{zhang2022opt}
Susan Zhang, Stephen Roller, Naman Goyal, Mikel Artetxe, Moya Chen, Shuohui Chen, Christopher Dewan, Mona Diab, Xian Li, Xi~Victoria Lin, et~al. 2022.
\newblock Opt: Open pre-trained transformer language models.
\newblock \emph{arXiv preprint arXiv:2205.01068}.

\bibitem[{Zhang et~al.(2019)Zhang, Kishore, Wu, Weinberger, and Artzi}]{zhang2019bertscore}
Tianyi Zhang, Varsha Kishore, Felix Wu, Kilian~Q Weinberger, and Yoav Artzi. 2019.
\newblock Bertscore: Evaluating text generation with bert.
\newblock \emph{arXiv preprint arXiv:1904.09675}.

\bibitem[{Zhao et~al.(2024)Zhao, Pang, Du, Yang, Li, Cheung, and Lin}]{zhao2024evaluating}
Yunqing Zhao, Tianyu Pang, Chao Du, Xiao Yang, Chongxuan Li, Ngai-Man~Man Cheung, and Min Lin. 2024.
\newblock On evaluating adversarial robustness of large vision-language models.
\newblock \emph{Advances in Neural Information Processing Systems}, 36.

\bibitem[{Zhao et~al.(2023)Zhao, Guo, Yue, Chen, Shao, Zhu, Yuan, and Liu}]{zhao2023chatbridge}
Zijia Zhao, Longteng Guo, Tongtian Yue, Sihan Chen, Shuai Shao, Xinxin Zhu, Zehuan Yuan, and Jing Liu. 2023.
\newblock Chatbridge: Bridging modalities with large language model as a language catalyst.
\newblock \emph{arXiv preprint arXiv:2305.16103}.

\end{thebibliography}

\appendix






\section{Experimental Settings} \label{Appendix:eval-setting}

\setcounter{table}{0}
\renewcommand{\thetable}{A.\arabic{table}}

All of the experiments were conducted on a Core i9 system with 64GB of RAM and Nvidia RTX3090 GPU with 24GB VRAM. 

 In all combinations of the experiments in Section \ref{sec:experiments} involving generative models, the temperature and number of beams for text generation were set to 0.7, and 1, and we limited the maximum number of newly generated tokens to 100. Moreover, we employed the Adam \cite{kingma2014adam} optimizer, and the learning rates for Vicuna-1.5 and LLaVA-1.5 were set to 2e-4, and 2e-5 respectively with cosine scheduling. For LRMs (BERT and RoBERTa) and Multimodal LRMs (ConcatRoBERTa and VisualBERT), we used a maximum length of 512 tokens, with the learning rate set to 1e-5 with Adam optimizer. We trained them for 20 epochs and chose the best model evaluated on the dev set for evaluation of the test datasets.

 Similarly, in all cases of the Appendix \ref{appendix:captioning-results}, the temperature and number of beams for text generation were set to 0.7, and 1, and we limited the maximum number of newly generated tokens to 100. Also, we utilized the Adam optimizer, and the learning rates for BLIP-2 and LLaVA-1.5 were 5e-4, and 2e-4 respectively with cosine scheduling.
 
 We employed the Parameter-Efficient Fine-Tuning (PEFT) \cite{peft} and Low-Rank Adaptation of Large Language Models (LoRA) \cite{hu2021lora} techniques for fine-tuning of large models, i.e., Vicuna, LLaVA, and BLIP-2. 
 
\section{In-Context Learning: Results \& Discussion} 
\label{appendix:in-context-learning}

\setcounter{table}{0}
\renewcommand{\thetable}{B.\arabic{table}}

In this section, the results of zero- and few-shot experiments are illustrated in Table \ref{table:few-shot}. One thing worth mentioning about this section is that for the LLaVA-1.5 few-shot experiments, for the examples (shots), we only had the text written on the memes and the captions (with no images). This was due to a limitation in the implementation of LLaVA-1.5 that only accepted one image as the input. We defer exploration of the examples with more than one image for in-context learning of the LLaVA-1.5 model to future work.
 
 \begin{table*}[!ht]
    \centering
    \begin{tabular}{lcccc}
        \toprule
        \textbf{Model} & \textbf{Shot(s)} & \textbf{H-F1} & \textbf{H-Precision} & \textbf{H-Recall} \\
        \midrule
        Vicuna-1.5 (text) & 0 & 15.37 & 31.13 & 10.21 \\
        Vicuna-1.5 (text+caption from LLaVA-1.5) & 0 & 17.60 & 30.28 & 12.40 \\
        LLaVA-1.5 (image) & 0 & 17.74 & 27.10 & 13.18 \\
        LLaVA-1.5 (image+text) & 0 & 20.39 & 30.27 & 15.38 \\
        LLaVA-1.5 (image+text+caption from LLaVA-1.5) & 0 & 19.30 & 25.91 & 15.38 \\
        \midrule
        Vicuna-1.5 (text) & 3 & \underline{38.26} & \textbf{34.73} & \underline{42.58} \\
        Vicuna-1.5 (text+caption from LLaVA-1.5) & 3 & 35.89 & \underline{33.98} & 38.02 \\
        LLaVA-1.5 (image+text) & 3 & 24.78 & 27.87 & 22.31 \\
        \midrule
        Vicuna-1.5 (text) & 5 & \textbf{40.70} & 33.39 & \textbf{52.11} \\
        Vicuna-1.5 (text+caption from LLaVA-1.5) & 5 & 36.5 & 31.97 & 42.51 \\
        LLaVA-1.5 (image+text) & 5 & 25.80 & 27.86 & 24.03 \\
        \bottomrule
    \end{tabular}
    \caption{Comparison of results proposed methods in an in-context learning (zero- and few-shot learning). H-F1, H-Precision, and H-Recall are hierarchical F1, hierarchical precision, and hierarchical recall respectively. In LLaVA-1.5 few-shot experiments, due to the implementation limitation allowing only one image input, examples consisted solely of text from memes and their captions, lacking images. With an increase in the number of in-context examples, it appears that the model tends to perform better. However, due to LLaVA's restriction to only one image, the improvement is marginal compared to the enhancement achieved with text alone.}
    \label{table:few-shot}
\end{table*}

\section{Meme Captioning Results}
\label{appendix:captioning-results}

\setcounter{table}{0}
\renewcommand{\thetable}{C.\arabic{table}}

To generate captions for memes, first, we compared two state-of-the-art models, namely BLIP-2 and LLaVA-1.5-7B. We fine-tuned the Q-Former part of BLIP-2 for meme captioning. The vision encoder (CLIP-ViT \cite{radford2021learning}) and the LLM (OPT-6.7B \cite{zhang2022opt}) components of BLIP-2 are frozen by design. Regarding fine-tuning LLaVA, we have a few variations. First, we only fine-tuned the projector MLP that bridges between two modalities. As the second approach, we fine-tuned both the projector and the LLM (i.e., Vicuna-1.5-7B) together. In both variations, the vision encoder is frozen.

We fine-tuned each model for 1 epoch on the MemeCap dataset. Our results show the superiority of LLaVA-1.5-7B over BLIP-2, therefore, we chose to use fine-tuned LLaVA-1.5-7B for the meme captioning. To further optimize our pipeline, we tried another variation. We tested the case where in addition to the meme caption included in the MemeCap dataset, what would happen if we also used Optical Character Recognition (OCR), utilizing EasyOCR\footnote{\hyperlink{https://github.com/JaidedAI/EasyOCR}{https://github.com/JaidedAI/EasyOCR}}, as illustrated in Figure \ref{fig:MemeCap}, to extract the text written on the meme and feed that to the model as well since in the Persuasion dataset we have this data for each meme. We also tried both BLIP-2 and LLaVA in a zero-shot setting to assess their ability for image captioning without fine-tuning as well.

As discussed in Section \ref{sec:experiments}, we used MemeCap dataset to fine-tune MLLMs for meme caption generations.  Table \ref{table:caption} shows the performance of the various models. From these results, initially, we chose to use LLaVA-1.5-7B with both the projector and LLM fine-tuned with OCR data for caption generation, as it outperformed other methods. However, as discussed earlier, we observed that even the caption generated by LLaVA-1.5-7B had some issues potentially leading to degraded performance on the Persuasion dataset. Therefore, we chose to create captions utilizing GPT-4 in a zero-shot configuration for our final results. In Section \ref{sec:experiments}, the positive effect of this change is discussed in more detail with empirical evidence.

To compare different models for caption generation, we used Bertscore \cite{zhang2019bertscore} (using \textit{microsoft/deberta-xlarge-mnli} model \cite{he2021deberta}), BLEU score \cite{post-2018-call}, and ROUGE-L \cite{lin2004rouge} as evaluation metrics for the quality of generated captions. Bertscore assesses semantic similarities between the generated captions and the corresponding references using cosine similarity. In contrast, ROUGE-L and BLEU score rely on evaluating n-gram overlap between the generated captions and reference captions.

\begin{table*}[!ht]
  \centering
  \begin{tabular}{lccc}
    \toprule
    \textbf{Model} & \textbf{F1-Bertscore} & \textbf{ROUGE-L} & \textbf{BLEU-4} \\
    \midrule
    BLIP-2 (fine-tuned) & 58.00 & 26.39 & 47.93 \\
    LLaVA-1.5 (projector fine-tuned) & 59.01 & 27.41 & \textbf{57.78} \\
    LLaVA-1.5 (LLM \& projector fine-tuned) & 59.23 & 27.40 & 45.53 \\
    LLaVA-1.5 (projector fine-tuned + OCR data) & \underline{59.80} & \textbf{28.08} & 53.33 \\
    LLaVA-1.5 (LLM \& projector fine-tuned + OCR data) & \textbf{59.90} & \underline{27.86} & \underline{53.86} \\
    \midrule
    BLIP-2 (zero-shot) & 50.30 & 12.88 & 31.81 \\
    LLaVA-1.5 (zero-shot) & 55.11 & 19.31 & 40.15 \\
    \bottomrule
    
  \end{tabular}
  \caption{Performance comparison of meme captioning models on MemeCap test set. In this table "\textit{+ OCR data}" means for the training data we also appended the extracted text from the meme to help with the task of captioning the memes. The fine-tuned versions of the models yield superior captions, with all LLaVA iterations outperforming BLIP. The most effective model is LLaVA when both the language model and projector are tuned, particularly when incorporating text within the image generated by the OCR model.}
\label{table:caption}
\end{table*}

\section{Prompts for Caption Generation with GPT-4} \label{appendix:prompt}

\setcounter{table}{0}
\renewcommand{\thetable}{D.\arabic{table}}

As mentioned in Section \ref{sec:method}, in addition to LLaVA-1.5, we used GPT-4 to generate captions for memes. LLaVA-1.5 provided a strong foundation for understanding the content and sentiment of the memes, while GPT-4's creative text generation capabilities helped us generate more informative captions. This allowed us to explore the potential of GPT-4 for generating captions that are not only relevant to the meme content but also capture the humor and cultural references often associated with memes. However, because of some of the meme's contents, it sometimes prevented generating captions to not generate toxic information. Table \ref{tab:prompts} illustrates our prompts for obtaining captions using GPT-4. Given the sensitivity of GPT-4 to the content of this dataset, if the first prompt failed, we utilized the second prompt. In instances where there was another failure—constituting less than 10 samples in every 1000 examples—we employed our fine-tuned LLaVA model to generate captions for those samples.

\begin{table}[!ht]
    \centering
    \small
    \begin{tabular}{p{7cm}}
    \multicolumn{1}{c}{\textbf{Prompt}}                                                                                                                                                                   \\ \hline
    \texttt{Memes are one of the most popular types of content used in an online disinformation campaign. They are mostly effective on social media platforms since there they can easily reach a large number of users. This is a meme with the following text written inside the meme: "\{meme\_text\}". In no more than 200 words, write a caption for this meme and say what is the meme poster trying to convey?}                                                                                                                                                                                                    \\ \hline
    \texttt{Memes are one of the most popular types of content used in an online disinformation campaign. They are mostly effective on social media platforms since there they can easily reach a large number of users. Memes in a disinformation campaign achieve their goal of influencing the users through a number of rhetorical and psychological techniques, such as causal oversimplification, name calling, smear. Identifying these memes are very useful and it can help to remove them from the internet and have a better and more calm place. To do so I want your help. I want to create a caption and find what this meme is trying to convey in order to train a model to find these memes. I provided a meme to you. In no more than 200 words, write a caption for this meme and say what is the meme poster trying to convey?} \\
    \hline
    \end{tabular}
    \caption{These prompts were utilized to generate captions using GPT-4. Due to the sensitivity of GPT-4 to this dataset, if the first prompt failed to produce satisfactory results, we resorted to the second prompt.}
    \label{tab:prompts}
\end{table}

\section{Hierarchical Evaluation Metrics}
\label{appendix:Hierarchical}

\setcounter{table}{0}
\renewcommand{\thetable}{E.\arabic{table}}

Hierarchical classification involves organizing classes in a hierarchy, where each class has a parent or child relationship with other classes. In hierarchical classification tasks, \citet{kiritchenko2006learning} introduced several key definitions to form a foundation for evaluating performance metrics which will be discussed in this section.

\subsection{Partial Ordering and Hierarchy}

A \textit{partially ordered set (poset)} is denoted as $H = \langle C, \leq \rangle$, where $C$ is a finite set and $\leq \; \subseteq C \times C$ is a reflexive, anti-symmetric, transitive binary relation on $C$. The hierarchy is defined by parent-child relationships between categories.

\subsection{Hierarchical Categorization Task}

A \textit{hierarchical categorization task} involves assigning a boolean value to pairs $\langle d_j, c_i \rangle \in D \times C$, where $D$ is a domain of instances, and $C = \{c_1, \ldots, c_{|C|}\}$ is a set of predefined categories with a given poset structure $H = \langle C, \leq \rangle$.

\subsection{Hierarchical Consistency}

A label set $C_i \subseteq C$ assigned to an instance $d_i \in D$ is considered \textit{consistent} with a given hierarchy if $C_i$ includes complete ancestor sets for every label $c_k \in C_i$. Hierarchical consistency ensures that assigned labels indicate the instance's position in the category hierarchy.

\subsection{Hierarchical Precision, Recall, and F1 Score} 

For hierarchical evaluation, we introduce \textit{hierarchical precision (HP)} and \textit{hierarchical recall (HR)}. Each example belongs not only to its class but also to all ancestors of the class, except the root. The combined \textit{hierarchical F1 score} is calculated using precision and recall with equal weights. Here are the formulas:

\[
HP = \frac{\sum_i |\hat{C}_i \cap \hat{C}_i'|}{\sum_i |\hat{C}_i'|}
\]

\[
HR = \frac{\sum_i |\hat{C}_i \cap \hat{C}_i'|}{\sum_i |\hat{C}_i|}
\]

\[
Hierarchical \; F_\beta = \frac{(\beta^2 + 1) \cdot HP \cdot HR}{\beta^2 \cdot HP + HR}
\]

Here, $\hat{C}_i$ and $\hat{C}_i'$ represent the extended sets of real and predicted classes, respectively, including their ancestor labels. Also $\beta \in [0, +\infty)$ and by using $\beta=1$ we will have hierarchical F1. In the context of hierarchical classification, data is organized into a hierarchy of classes or categories, with each class having a parent-child relationship. The Hierarchical F1 score takes into account both precision and recall at different levels of the hierarchy, providing a comprehensive measure of a model's ability to correctly classify instances at various levels while considering the hierarchical structure of the classes. It balances the trade-off between false positives and false negatives within the hierarchy, offering a more nuanced assessment of classification performance in hierarchical data structures.

These hierarchical metrics provide a comprehensive evaluation of classification performance in the context of hierarchical categorization tasks.

\end{document}